\documentclass{goodfire}
\usepackage{goodfire}
\usepackage{import}

\title{The Shape of Beliefs: \\ Geometry, Dynamics, and Interventions \\ along Representation Manifolds of Language Models' Posteriors}

\newlength{\imgwidth}
\AtBeginDocument{\setlength{\imgwidth}{0.8\textwidth}}

\usepackage{microtype}
\usepackage{hyperref}
\usepackage{url}
\usepackage{booktabs}
\usepackage{graphicx}
\usepackage{wrapfig}

\usepackage{lineno}

\definecolor{darkblue}{rgb}{0, 0, 0.5}
\hypersetup{colorlinks=true, citecolor=darkblue, linkcolor=darkblue, urlcolor=darkblue}

\usepackage[percent]{overpic}
\usepackage{subcaption}
\usepackage{enumitem}
\usepackage{amsmath,amssymb}
\usepackage{stmaryrd}
\usepackage{xcolor}
\renewcommand{\vec}[1]{\boldsymbol{#1}} 

\authors{Rapha\"{e}l Sarfati\textcolor{goodfire2}{$^\ast$}, 
Eric Bigelow,
Daniel Wurgaft,
Siddharth Boppana,
Jack Merullo, \\
Atticus Geiger,
Owen Lewis,
Tom McGrath,
\& Ekdeep Singh Lubana\textcolor{goodfire2}{$^\ast$}
}

\abstract{
Large language models (LLMs) form implicit beliefs (posteriors over latent variables) from prompts, but we lack a mechanistic account of how these beliefs are encoded in representation space, how they update with new evidence, and how interventions reshape them. 
We study a controlled setting in which \texttt{Llama-3.2} infers the parameters of a normal distribution from in-context samples. 
We show that parameter posteriors are encoded as curved manifolds in representation space, and trace how they evolve along the prompt.
Standard linear steering moves representations off-manifold, inducing unintended, coupled changes, whereas geometry-aware methods preserve the target belief family.
Our work demonstrates an example of linear field probing (LFP) as a principled approach to tile the data manifold and make interventions that respect the underlying geometry. Our results suggest that LLM beliefs are inherently geometric objects, and that globally linear representations are often inadequate abstractions.
Code available at \href{https://github.com/raphael-goodfire/shape-of-beliefs}{raphael-goodfire/shape-of-beliefs}.
}

\begin{document}



\maketitlebox


\section{Introduction}

Natural language is inherently ambiguous: the same surface form can express multiple meanings, roles, or functions, resolved only through context. 
For instance, the English suffix ``-s'' can mark plurality (``trees''), possession (``Dylan's''), or verb agreement (``runs'')~\citep{piantadosi2012communicative}. 
As a result, any system that understands and acts through language must engage with this ambiguity: it must represent uncertainty over latent states (intended meanings, referents, world states, task hypotheses) consistent with the input, and update that uncertainty as new evidence arrives. 


\begin{figure}[h!]
\centering
\begin{overpic}[width=1.00\linewidth]{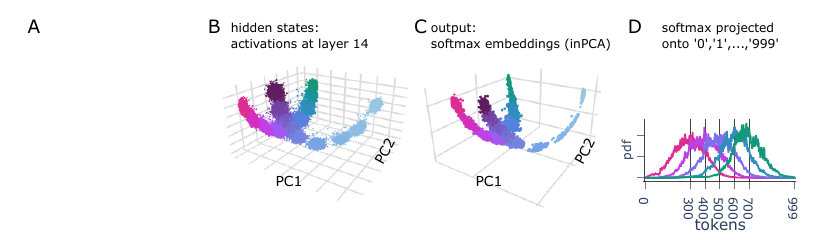}
    \put(0,5.5){\includegraphics[width=0.25\linewidth]{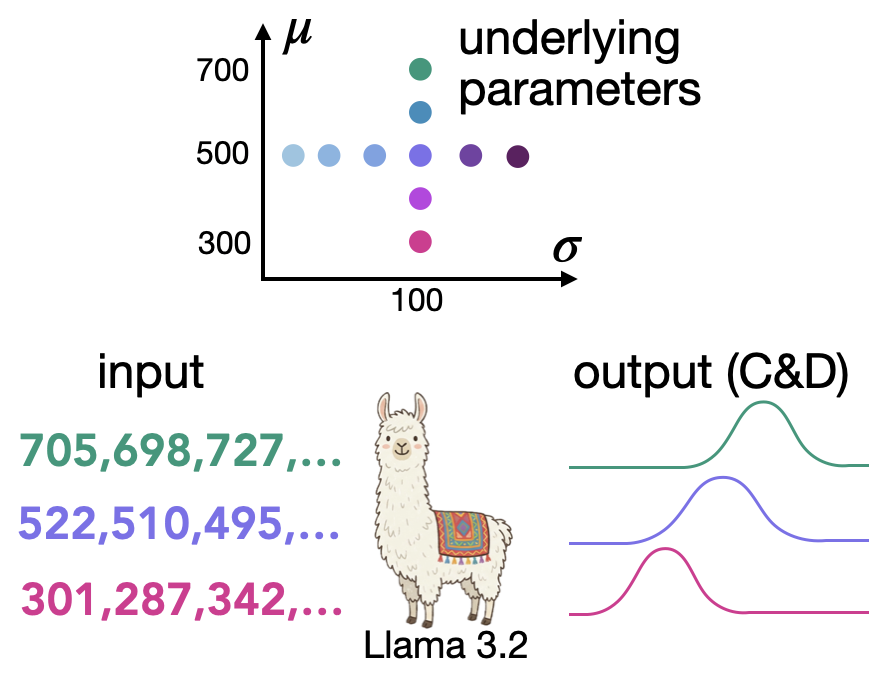}}
\end{overpic}
\caption{
  \textbf{Shape of beliefs.}
  (A)~Stochastic time series $u(t) \in \llbracket 0, 999 \rrbracket$ are generated from normal distributions $\mathcal{N}(\mu,\sigma)$ draws and passed into \texttt{Llama~3.2} as strings.
  (B)~Representations (PCA at layer 14) form curved manifolds parametrized by $(\mu, \sigma_0=100)$ (pink to green) and $(\mu_0=500,\sigma)$ (shades of purple).
  They encode the model's current posterior inferred from the input data.
  (C) Softmax outputs, shown here in as a 3D embedding representation from inPCA, mirror the geometry of activations.
  (D) Softmax probabilities, alternatively represented by their projection onto the subset of tokens corresponding to integers from 0 to 999, closely reflect the input distributions.
}
\label{fig:belief_shape}
\end{figure}

Large language models instantiate the requirement above in a concrete way: at each position, next-token prediction is driven by an internal state that compresses a distribution over the plausible latent structures behind the prompt---i.e., an implicit belief state. Studying and altering model behavior can then be boiled down to the questions of assessing how a model encodes beliefs over latent variables of a distribution, how these beliefs are updated in light of new evidence, and how beliefs can be altered by intervening on model representations.

\paragraph{This work.} We study the questions above in a toy task that makes tractable our core objects of interest: beliefs over latent concepts with known ground truth. Building on recent in-context learning work showing that models can infer latent variables of a data-generating process from examples in the prompt and reflect those variables in their predictive distributions~\citep{bigelow2025beliefdynamicsrevealdual, liu2025density}, we feed the model sequences of comma-delimited integers sampled from a generating distribution $\mathcal{G}_{\vec{\theta}}$. The model converges to a next-token prediction that closely matches $\mathcal{G}_{\vec{\theta}}$~\citep{liu2025density}, giving us a controlled notion of a posterior over a small set of interpretable parameters $\vec{\theta}$ (e.g., $\vec{\theta}=(\mu,\sigma)$ for Gaussians) that lets us measure both outputs (logits/softmax) and their preimages (hidden states) in a calibrated way with ground-truth anchors.
From this controlled setup, we address the following questions. 

\begin{itemize}
    \item \textbf{Geometry:} How is a prompt-conditioned posterior belief encoded in the model's internal representations, and what geometric structure does it form? We characterize the geometry of posterior belief in representation space by showing \emph{smooth but curved} manifolds associated with different inferred beliefs over $(\mu,\sigma)$. We find these belief states can be read out via a collection of linear probes that ``tiles'' the manifold (termed linear field probes by \citet{yocum2025neural}).    
    \item \textbf{Dynamics:} When evidence changes over time, how does the model update that posterior---do trajectories move along or across belief manifolds? To answer this question, we consider sequences where the generating distribution switches mid-sequence and quantify the resulting belief-update trajectories and the characteristic timescales by which the model equilibrates.     
    \item \textbf{Interventions:} If we try to manipulate a belief by intervening on activations, when do standard steering methods succeed~\citep{marks2024the, rimsky2024contastive}, and when do they produce unintended behavior? We compare standard linear steering to geometry-aware approaches, showing that interventions which respect manifold structure better preserve the intended belief family and avoid unintended coupled shifts.
\end{itemize}

Overall, our work suggests that beliefs in LLMs are best understood as structured, geometric objects in representation space, not as single linear directions. 
We argue these representation manifolds can serve as building blocks for more reliable interpretability and intervention—enabling edits that target the intended belief update while preserving the surrounding family of behaviors.

\section{Experiments and analysis}
Our goal is to make a posterior belief state both observable in the logits and indexable by a small set of interpretable parameters, so we can study its preimage geometry and update trajectories along this manifold.
Stochastic numerical time series provide this setting: the model's next-token distribution can be compared directly to the ground-truth generating distribution~$\mathcal{G}(\theta)$, and varying $\theta$ provides a continuous family of belief states. 
In particular, we consider the setup by \citet{liu-etal-2024-llms-learn}, who showed that LLMs are capable of various time series extrapolation tasks; in particular, given a prompt such as \texttt{533,460,689,432,501,487,508,465,340, $\dots$}, where the numbers are random variables drawn from a normal distribution~$\mathcal{N} \left(\mu=500,\sigma=100 \right)$ (as an example), the model quickly in-context learns the distribution underlying the input series of (stringified) numbers, and reproduces this distribution in its logits (App.~\ref{app:convergence}).
We rely primarily on the \texttt{Llama-3.2-1B} model~\citep{grattafiori2024llama3herdmodels}, whose tokenizer discretely represents every single number between 0 and 999~\citep{bao2025texttrainedllmszeroshotextrapolate}, and restrict input distributions to integers in that range by rounding and clamping.
We extend the analysis to several other models in~App.~\ref{app:models}.
In our setup, there are two categories of tokens: the comma-predicting-number (\texttt{com2num}) tokens and the number-predicting-comma (\texttt{num2com}) tokens. 
Unless otherwise noted, we consider the \texttt{com2num} tokens in what follows.
More details in App.~\ref{app:convergence}.

\paragraph{}
The code to generate data, analyze data, and reproduce figures is made available at:
\newline
\href{https://github.com/raphael-goodfire/shape-of-beliefs}{https://github.com/raphael-goodfire/shape-of-beliefs}.

\subsection{Methods}
We use Principal Component Analysis (PCA) to visualize and analyze high-dimensional sets of vectors.
For vectors describing probability distributions, such as softmax outputs, we employ \emph{Intensive PCA} (inPCA) instead, a variation of PCA more informative for prediction vectors, i.e., vectors constrained to the unit $d$-simplex~\citep{doi:10.1073/pnas.1817218116}.

\subsection{Notations and definitions}

We write vectors in bold (e.g.,~$\vec{v}$) and matrices in capital letters (e.g.,~$\vec{A}$).
We use \textit{activations}, \textit{representations}, and \textit{hidden states} interchangeably to refer to the model's residual stream, denoted~$\vec{x}_{i,k}(l)$, where $i$~is the token index, $k$~the sequence index, and $l \geq 0$~the layer index.
A normal distribution with mean~$\mu$ and standard deviation~$\sigma$ is written~$\mathcal{N}(\mu,\sigma)$ or~$\mathcal{N}_{\mu,\sigma}$.
Its \emph{preimage} manifold is denoted $\mathcal{M}_{\mu,\sigma}$: if $\vec{x} \in \mathcal{M}_{\mu,\sigma}$, after layer-norm and unembedding it maps to a softmax corresponding to~$\mathcal{N}_{\mu,\sigma}$.
For preimages at intermediate layers~$l$ (not the final layer), we write~$\mathcal{M}_{\mu,\sigma}^{(l)}$.

We use logits as a proxy for model behavior.
To compare input distributions to output behavior, we normalize the logits into probability vectors by applying softmax with temperature $T=1.0$.
Finally, we often use the term ``Linear Field Probes'', as proposed by \citet{yocum2025neural}, for describing a family of linear probes that are trained to be sensitive to a specific part of the input domain (e.g., sensitive to specific values of $\mu, \sigma$). 
These probes help us elicit an intriguing duality between beliefs, as observed in the output space, and the geometry of representations that encodes these beliefs.

\section{Results}
We first investigate the geometry of belief manifolds tiled by the activations from various input time series.
Then, we report the dynamics along and across these manifolds when time series switch between distributions.
Finally, we demonstrate that linear field probes tile the manifold without privileged access to knowledge about the underlying geometry, and we use them to design causal interventions based on the discovered geometry.

\subsection{Shape of beliefs}
\label{sec:shape_of_beliefs}

Given enough number samples, of the order of $100$, \texttt{Llama~3.2} in-context learns the distribution and converges towards the true distribution underlying the data (App.~\ref{app:convergence}), as evidenced by its logit output in Fig.~\ref{fig:belief_shape}C \& D.
In other words, over the course of the prompt the model acquires a posterior belief about the input contextual information.
Critically, this belief state is not only manifested in the logits, but also encoded in the internal representations of the prompt. 
We extract an ensemble of activations corresponding to the preimages for normal distributions with various means and standard deviations~$\mathcal{N}_{\mu,\sigma}$.
Fig.~\ref{fig:belief_shape}B reveals two orthogonal manifolds, one corresponding to varying means $\mu \in \{300, 350, \ldots, 700\}$ and constant standard deviation $\sigma_0=100$, the other to constant mean $\mu_0 = 500$ and varying $\sigma \in \{20, 50, \ldots, 200\}$.
These manifolds appear to form smooth, continuous structures, 
exhibiting substantial curvature.
This implies, \textit{a priori}, that the geometry supporting the data is complex, and might not be adequately described by standard linear frameworks such as the Linear Representation Hypothesis (LRH, \citep{elhage2022superposition, 10.5555/3692070.3693675}).
These structured manifolds appear across models, as shown in App.~\ref{app:models}.

\begin{figure}
  \includegraphics[width=\columnwidth]{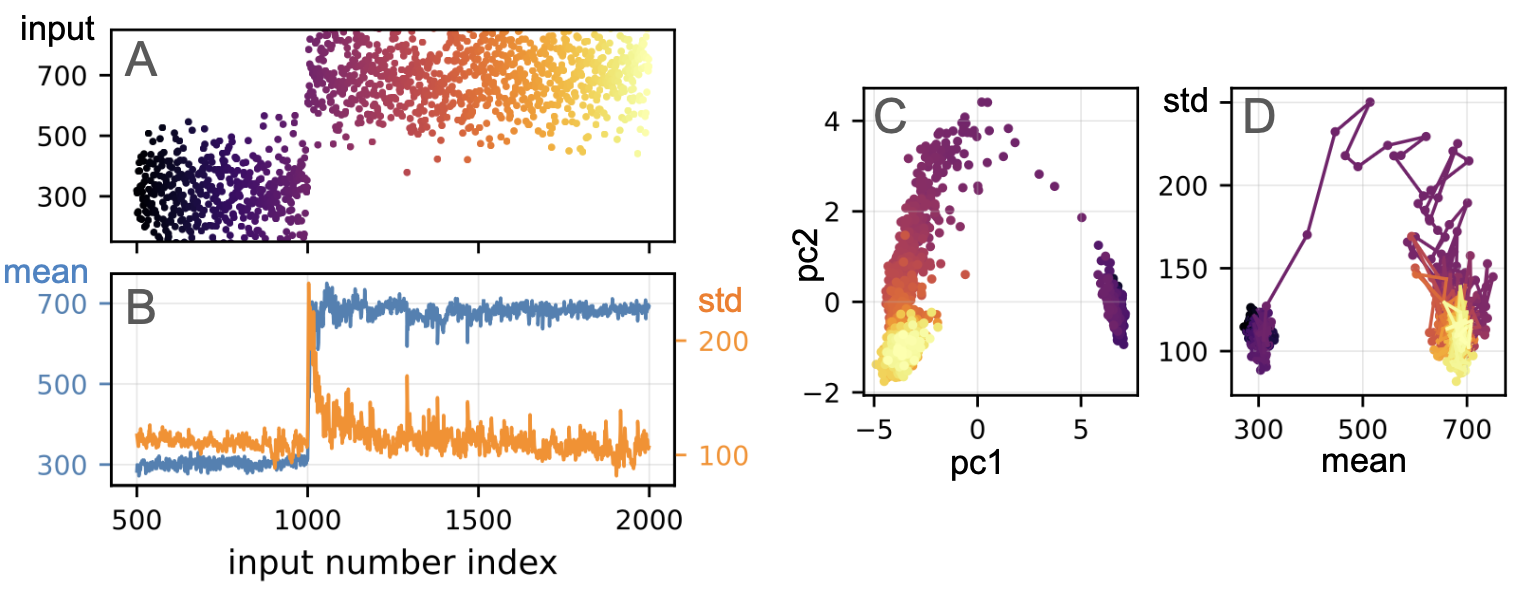}
  \caption{ 
  \textbf{Belief dynamics.}
  Input (A) is a time series with a sharp transition between two distinct regimes, $\mathcal{N}_{300,\sigma_0} \rightarrow \mathcal{N}_{700,\sigma_0}$ at $t=1000$.
  Output (B) shows the model quickly adapting its mean after the switch while broadening its variance, which relaxes back to its true value after about 300 tokens.
  This switch in belief is apparent in the model's activations (C).
  The trajectory in probability space (D) shows two attractors corresponding to the true input distributions, and the path taken by the model across them.
  }
  \label{fig:belief_dynamics}
\end{figure}

\subsection{Belief dynamics}
\label{sec:belief_dynamics}

Now that we have established a ground-truth map of beliefs, we investigate the situation where the input materializes a sharp change of distribution and the corresponding response of the model to this impulse perturbation.
Concretely, we study the model's output distribution when we input a time series whose first 1000 numbers follow $\mathcal{N}(300,100)$ and the next ones abruptly switch to $\mathcal{N}(700,100)$ (Fig.~\ref{fig:belief_dynamics}A).

We observe two timescales of adjustment in Fig.~\ref{fig:belief_dynamics}B. 
The mean of the output distribution quickly settles on the new mean. 
The variance takes longer to equilibrate, defining an effective \emph{timescale of belief equilibration}. 
The specific trajectory that the model follows is reflected both in probability space (Fig.~\ref{fig:belief_dynamics}D) and in representation space (Fig.~\ref{fig:belief_dynamics}C).
They show that, after the switch, the model jumps to a phase of high variance, or equivalently a phase of \emph{high entropy}\footnote{For a normal distribution, entropy $h = \log{\sigma} + h_0$.}, reflecting the model's uncertainty about what the input data represents.

In App.~\ref{sec:ideal_observer}, we compare the model's empirical response to that of an ideal observer.
In App.~\ref{app:metalearning}, we consider the response to a time series with multiple switches and observe meta-in-context learning capabilities~\citep{codaforno2023metaincontextlearninglargelanguage}.

\subsection{Features encoded along data manifolds}

With belief states mapped as manifolds and updates measured as trajectories, we investigate which coordinates of belief are accessible to simple readouts. 
This matters for two reasons: 
(i) probing: how we extract what the model believes without relying on prompting; 
and (ii) intervention: how we define a
principled steering objective, such as ``increase $\mu$ while preserving $\sigma$''. 
We therefore study how the $\mu$-parameter is decodable along the manifold using linear field probes.
 
\subsubsection{Linear field probes}
\label{sec:lfp}

To address the mounting evidence that neural networks, and transformers in particular, encode features using geometries and topologies far more complex than single directions~\citep{gurnee2025when}, \citet{yocum2025neural} recently introduced \textit{feature fields} to describe features defined over (non-Euclidean) manifolds---one can deem these fields as describing a ``higher-level concept'', e.g., the number system, compared to a ``lower-level concept'', e.g., a unique number.
Overall, the framework proposes a natural extension of linear probes into families of probes, called \textit{linear field probes} (LFP), which essentially operate a piecewise-linear tiling of an underlying manifold.

Using our continuous representation of beliefs, we examine the linear field probing on the gaussian-preimage manifolds $\mathcal{M}_{\mu,\sigma}^{(l)}$.
We find that linear field probes produce an excellent representation of the belief manifolds. 

\begin{figure}
\includegraphics[width=\columnwidth]{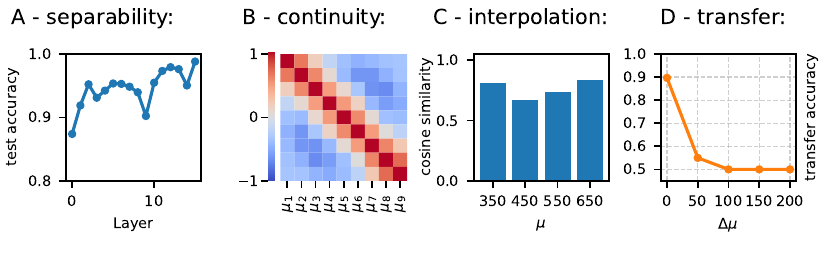}
  \caption{ 
  \textbf{Linear field probes.} 
  (A) Separability: probe accuracy (on test set) on all $\mu$-indexed representations, for all layer.
  (B) Continuity: cosine similarity between probes, showing smooth variation over $\mu$ and revealing structured geometry over the domain.
  (C) Interpolation: cosine similarity between true and (kernel-) interpolated (see~\ref{sec:probe_interpolation}) vectors at intermediate $\mu$; for reference, cosine similarity between two random vectors in a space of $d=2048$ dimensions has mean 0 and standard deviation $1/\sqrt{2048} \simeq 0.02$.
  (D) Transfer: probes only transfer locally; here showing a probe trained on $\mu = \{ 300, 350 \}$ (layer 0) and applied on sets $\mu = \{ 350, 400 \} (\Delta\mu =50)$, $\mu = \{ 400, 450 \} (\Delta\mu = 100)$, etc. (An untrained binary probe has accuracy~0.5.)
  }
  \label{fig:LFP}
\end{figure}

Specifically, we train a set of linear probes on activations classifying the discretized manifold $\sigma_0=100, \mu_i = \lbrace 300, 350, \dots, 700 \rbrace$, at every layer.
At each layer, we fit a single multiclass probe to predict $\mu$ via a softmax classifier trained with cross-entropy, using a 80/20 train-test split (see~App.~\ref{app:linearprobes} for more details). 
We report the following findings.
\begin{itemize}[leftmargin=12pt, itemsep=2pt, topsep=2pt, parsep=2pt, partopsep=2pt]
    \item \textbf{Separability}: probes achieve high accuracy on all activation classes $\mu_i$, from 0.87 at layer~0 to 0.99 at layer~15 (Fig.~\ref{fig:LFP}A); this shows that $\mu$-indexed representations are linearly separable at all layers.
    \item \textbf{Continuity}: probes vary smoothly with the parameter $\mu$, as revealed by the structured cosine similarity matrix in Fig.~\ref{fig:LFP}B; this smooth similarity structure reflects an underlying geometry over the domain $\mu$, as further discussed below.
    \item \textbf{Interpolation}: probe vectors can be interpolated across $\mu$ based on trained endpoints (Fig.~\ref{fig:LFP}C); for example, the probe at $\mu=350$ can be interpolated between $\vec{w}_{300}$ and $\vec{w}_{400}$ without retraining. This indicates that the probes form a coherent field over $\mu$ rather than a collection of unrelated classifiers.
    \item \textbf{Transfer}: probes transfer \emph{only locally} along the manifold; transfer performance decays with distance in $\mu$ at a rate matching the decay lengthscale in the Gram matrix (Fig.~\ref{fig:LFP}D). This indicates curvature in the domain embedding and limits the validity of a single (global) linear direction.
\end{itemize}

Together, these observations support the interpretation that the set of linear probes constitute a LFP as defined in~\citet{yocum2025neural}.
Separability establishes that $\mu$ is linearly decodable, while continuity and interpolation establish that $\mu$ is linearly represented as a field.
Thus, the manifold of gaussian activations is a feature field that is linearly represented.

\subsubsection{Field geometry} 

We stress that, in general, a standard multiclass probe for (non-parametric) classes~\citep{alain2018understandingintermediatelayersusing}, for example $\{$cats, dogs, horses, raccoons$\}$ would \emph{not} show the continuous and interpolative structure that characterizes linear field probes---because there is no semantic or conceptual continuity between these discrete categories.
A LFP, instead, represents a bilinear form $f(\vec{x},\mu) = \langle \vec{x},\Psi(\mu) \rangle$, where $\Psi(\mu)$ denotes the probe vector associated with field value $\mu \in \mathcal{Z}$; evaluation of the field is simply an inner product~$\langle \cdot, \cdot \rangle$, allowing to encode a space's structure in operand ($\Psi(.)$) and read said structure out from the other operand ($x$).
Correspondingly, the linear field probe implicitly endows the domain $\mathcal{Z} = [300,700]$ with a particular geometry.
Importantly, this geometry does not describe where activations corresponding to different domain values are (co-)located in activation space.
Instead, it describes which linear directions in activation space matter for decision boundaries:
the way $\mu$ is encoded is not the same as the way $\mu$ can be read out.
This distinction between encoding geometry and separability relates to the notion of primal vs.\ dual space in information geometry; see App.~\ref{app:dual} for further discussion of this subtle point.

\begin{wrapfigure}[18]{l}{0.5\textwidth}
  \includegraphics[scale=1]{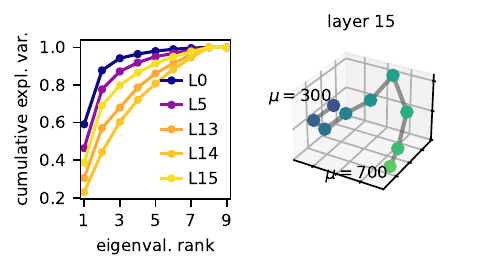}
  \caption{ 
  \textbf{Field geometry.} 
  (Left) Cumulative variance explained as a function of rank of the eigenvalues of the LFP Gram matrices (Fig~\ref{fig:LFP}B), for each layer. The intrinsic dimensionality increases with layers, but drops at the last layer (L15).
  (Right) Kernel PCA embedding of the first 3 eigenvectors of the Gram matrix at layer~15. This represents the field geometry, dual to the activation space manifold.
  }
  \label{fig:dual}
\end{wrapfigure}

\paragraph{}
This global structure of the linear readout space is naturally captured by the Gram matrix of probe vectors, defined via cosine similarity and shown in Fig.~\ref{fig:LFP}B.
The Gram matrix encodes which directions in activation space are functionally similar for readout, independently of where the data lies in the manifold.
Its eigendecomposition reflects the underlying dimensionality and directionality necessary to decode $\mu$ values from the embedded representations. 
Following standard kernel PCA, this dual geometry can be visualized by embedding each $\mu_i$ as
\begin{equation*}
    \left( \sqrt{\lambda_1} \vec{u}_1(\mu_i), \sqrt{\lambda_2} \vec{u}_2(\mu_i), \sqrt{\lambda_3} \vec{u}_3(\mu_i) \right), 
\end{equation*}
where $\lambda_\beta$ and $\vec{u}_\beta$ are the leading eigenvalues and eigenvectors of the Gram matrix (Fig~\ref{fig:dual}B).

Figure~\ref{fig:dual} thus illustrates a representation in which the readout geometry is dominated by a small number of principal directions, even when the underlying activation manifold is highly curved. 
This highlights the duality between encoding complexity and linear usability: \textit{complex internal representations can give rise to comparatively simple linear fields}.

\paragraph{}
Finally, the LFP also reveals the evolution of the field geometry across layers.
In particular, eigenvalues of the Gram matrix show that the intrinsic dimensionality of the manifold increases over layers (Fig.~\ref{fig:dual}A)---with the exception of the last layer\footnote{This ``last layer anomaly''~\citep{sarfati2025lines} has been noted many times before, but remains to this day to be characterized and explained precisely. Intuitively, the model converges to a next token prediction.}. 
It may be interpreted as the deeper representations encoding an increasing number of features, requiring the feature field to densify and spread across more dimensions in readout space.

\section{Interventions}

Based on the model's belief map, can we purposefully intervene on its activations and engineer its output to match a specific distribution?
This is traditionally referred to as \emph{model steering}~\citep{subramani2022extracting, turner2024steeringlanguagemodelsactivation}.

Model steering is usually performed along a specific direction with a tunable magnitude; in other words: \emph{linearly}. 
However, the manifold geometry and the field probe results uncovered above suggest that steering ought to be done along the manifold that supports the underlying structure rather than a constant direction \citep{hindupur2025projecting, lubana2025priors, fel2025rabbithulltaskrelevantconcepts}. 
Here, we use our experiments to compare various steering schemes and help concretize this point precisely. 
In summary, we find that steering linearly tends to push next-token prediction ``off the data manifold'', yielding predictions that are misaligned with the target distribution.

\subsection{Activation-aware steering (primal space)}
Assume we have an ensemble of last-layer representations $\vec{x}_{i, k}^{(300)} \in \mathcal{M}_{300, 100}$ mapping onto next-token predictions $\sim \mathcal{N}_{300, 100}$, 
and a second set $\vec{x}_{i, k}^{(700)} \in \mathcal{M}_{700, 100}$ mapping onto $\mathcal{N}_{700, 100}$.
Suppose we aim to steer the $\mu=300$~representations towards an intermediate target $\mathcal{N}(500,100)$. 
A simple steering approach computes the ``difference of means'' steering vector
\begin{equation*}
    \vec{s} = \overline{\vec{x}}_{700} - \overline{\vec{x}}_{300},
\end{equation*} 
where $\overline{\vec{x}}_{300}, \overline{\vec{x}}_{700}$ are class centroids, and applies the linear intervention::
\begin{equation*}
\tilde{\vec{x}}_{i, k}^{(500)} \simeq \vec{x}_{i, k}^{(300)} + \alpha \vec{s},    
\end{equation*}
with $\alpha = 0.5$ (halfway between 300 and 700). 
This is essentially Contrastive Activation Steering~\citep{rimsky-etal-2024-steering}.

\begin{figure}
\centering
\begin{subfigure}{0.6\linewidth}
\centering
\includegraphics[width=\linewidth]{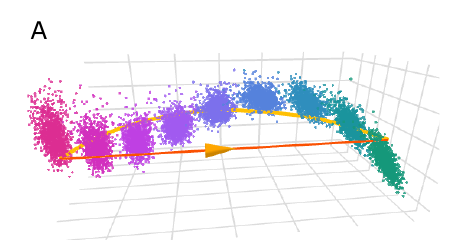}
\end{subfigure}
\hfill
\begin{subfigure}{0.39\linewidth}
\centering
\includegraphics[width=\linewidth]{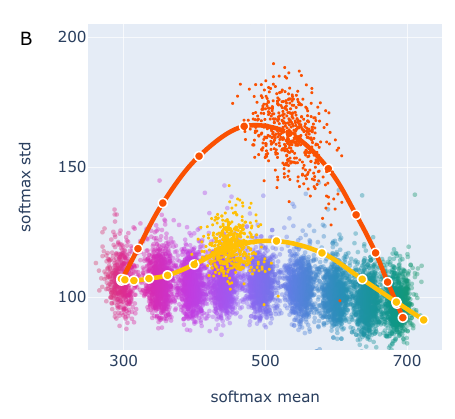}
\end{subfigure}
\caption{
  \textbf{Steering based on activation geometry: linear vs manifold-aware.}
  The manifold of activations (A) for $\mu \in [300, 700]$ provides a principled basis for steering directions, for example a centroid-to-centroid vector (orange), and a manifold-fitting spline (yellow).
  The corresponding resulting logits are shown in (B). 
  Steering linearly (orange path) brings steered logits far out of the $\sigma=100$-distributions, while steering \emph{along} the manifold keeps outputs significantly more aligned with the targeted behavior.
}
\label{fig:steering_primal}
\end{figure}

However, the set $\mathcal{M}_{\mu,100}$ of the preimages to $\mathcal{N}_{\mu,100}$ varies nonlinearly with $\mu$, and the resulting trajectory is curved in activation space (Fig~\ref{fig:steering_primal}A).
Thus, steering linearly towards $\mathcal{M}_{700,100}$ might push activations out of the gaussian manifold, so that when they reach $\mu=500$ their shape departs significantly from that of the normal distribution $\mathcal{N}_{500,100}$.
This is indeed what we observe experimentally in Fig.~\ref{fig:steering_primal}B, where the linear steering in activation space forms a curve in logit (image) space so that steered activations reaching $\mu=500$ have a large variance (orange curve).
In contrast, if we parametrize along the prototypes of $\mathcal{M}_{\mu,\sigma_0=100}$ and steer \emph{along the manifold}, we preserve the class-conditioned structure: induced predictions remain close to $\mathcal{N}(\mu, \sigma_0)$ while shifting their mean towards $\mu=500$ (Fig.~\ref{fig:steering_primal}B yellow curve).
These results thus offer concrete evidence that linear steering can lead to unexpected and unpredictable behaviors when the underlying geometry is curved, indicating one ought to first characterize the geometry before steering is performed.

\subsection{Field-aware steering (dual space)}
A complementary approach to intervening in the representation space is to steer the model in the \emph{dual} (i.e., the distribution) space using probe vectors~\citep{Li2023,10903266,park2026informationgeometrysoftmaxprobing}.
Specifically, rather than steering \emph{towards activation prototypes}, this method attempts to modify the component of the representation that controls a linear readout of~$\mu$.

\paragraph{Linear probe steering.}
In our setup, we can use point probes to define the steering vector: 
\begin{equation*}
\vec{s}^\star = \frac{\vec{w}_{\mu=700} - \vec{w}_{\mu=300}}{\lVert \vec{w}_{\mu=700} - \vec{w}_{\mu=300} \rVert}.
\end{equation*}
Steering $\vec{x}^{(300)} \in \mathcal{M}_{300}$ towards $\mathcal{M}_{700}$ reads:
\begin{equation*}
    \vec{x}_s(\alpha) = \vec{x}^{(300)} + \alpha \vec{s}^\star.
\end{equation*}
One immediate caveat is that, since \emph{feature geometry is agnostic to scale}, it's unclear which values of $\alpha$ are meaningful.
Prior work has found that small $\alpha$ can have negligible effects on logits (\citet{heimersheim_mendel_2024_activation_plateaus}'s ``activation plateaus''), whereas large values break the model towards distribution shifts and degenerate outputs.
We directly visualize the evolution of the induced output distribution as a function of~$\alpha$ in Fig.~\ref{fig:dual}.
Again, this naive linear steering scheme pushes output distributions away from the Gaussian manifold.

\paragraph{Field-aware steering from LFP.}
Alternatively, the linear field probe provides a geometry over $\mu$ to extract field-aware steering vectors.

\begin{wrapfigure}{l}{0.5\textwidth}
\centering
  \includegraphics[scale=0.8]{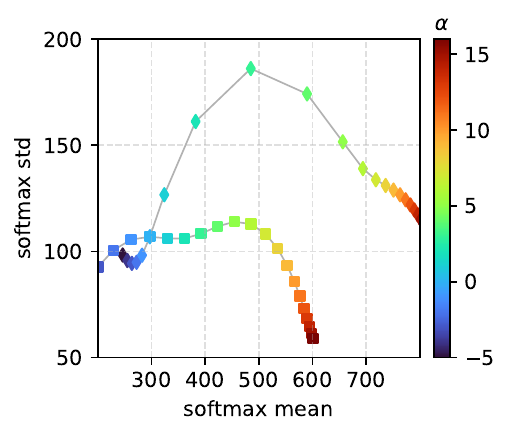}
  \caption{ 
  \textbf{Steering based on feature and field geometry.}
  Steering along a single probe direction, here $\vec{w}_{300 \rightarrow 700}$ (lozenges $\blacklozenge$) again brings the steered activations far from the $\sigma_0 = 100$ manifold.
  Taking advantage of the linear field probe to smoothly travel along the field geometry (squares $\blacksquare$), however, maintain the same standard deviation while increasing the mean -- at least in a certain regime between 300 and 500.
  }
  \label{fig:steering_dual}
\end{wrapfigure}
\paragraph{} 
Specific implementations vary; we propose the following.
We use the first $r$ eigenmodes of the LFP Gram matrix~$\vec{K}$ and obtain a low-dimensional embedding:
\begin{equation*}
c(\mu) = \left( \sqrt{\lambda_1}\vec{u}_1 (\mu), \dots, \sqrt{\lambda_r}\vec{u}_r (\mu) \right).  
\end{equation*} 
We then fit a spline $\tilde{\vec{c}}(\mu)$ and use it to define an interpolated steering direction
 $\vec{s}^\star = \sum_i \alpha(\mu) \vec{w}_{\mu_i}$, where the weights $\alpha_i$ are obtained by kernel regression using~$\vec{K}$.
This removes noisy components and preserves only the smooth and predictive geometry.
As shown in Fig.~\ref{fig:steering_dual}, this approach maintains the induced output distribution closer to the intended $\mathcal{N}_{\mu,\sigma_0}$ family, within a certain range.

\section{Conclusion}
\label{sec:conclusion}
We introduced a principled framework for tracking language models' beliefs by mapping how posterior predictions are encoded as structured manifolds in activation space, and how non-stationary inputs induce dynamics between attractor regions.
Using this setup, our main finding is that purely linear concept representations are often an inadequate abstraction. 
Even when a target feature, such as the mean of an output distribution, is locally decodable, the underlying representations can exhibit substantial curvature.
Moreover, we show in App.~\ref{app:mixture} that the joint $(\mu,\sigma)$ sheet cannot be reconstructed as a simple additive composition of the anchor manifolds $\mathcal M_{\mu,\sigma_0}$ and $\mathcal M_{\mu_0,\sigma}$, suggesting non-linear interactions between belief parameters.
Hence, linear steering can move activations off-manifold, producing unintended coupled shifts in other covariates (e.g., distort the shape of the distribution).
This geometric mismatch reflects the gap between readout geometry and representation geometry.
Linear separability does not entail geometrically faithful control: \textbf{features may be linearly separable and readable, yet linear steering can still cut across (rather than along) the curved belief manifold}.
This is notably the case when the probe field spans multiple directions ($\mathrm{rank} > 1$; see App.~\ref{app:dual}).

Our work is part of a broader effort to bridge representation geometry and intervention schemes in language models. 
Increasingly, representations are understood beyond collections of linear features, as structured geometric objects shaped by statistical correlations and symmetries in the training data~\citep{karkada2026symmetrylanguagestatisticsshapes,prieto2026datastatisticsfeaturegeometry}. This perspective is formalized in recent work on the information geometry of softmax representations: \citet{park2026informationgeometrysoftmaxprobing} propose that the natural notion of distance is not Euclidean but induced by the divergence between output distributions, leading to a dually flat geometry with distinct primal and dual structures. 
This geometric lens motivates alternatives that operate in the appropriate coordinate system; for example, by projecting activations back onto learned data manifolds~\citep{luo2026learninggenerativemetamodelllm} or by gradient-informed paths between contrastive clusters~\citep{zhao2026odesteerunifiedodebasedsteering}. 
Our findings complement this view: effective control of belief systems, and hence behavior, must respect the intrinsic geometry of representations.

\paragraph{Limitations.} Our analysis relies on true activations of base models without finetuning; nevertheless, it is currently constrained to numerical settings.
Extending these findings to broader natural language processing contexts remains open, particularly because identifying continuous parametrizations in natural language is more challenging. 
However, analogous continuous and quantitative descriptions of other rather intangible objects, such as stories~\citep{Reagan2016EmotionalArcs} or behavior~\citep{Stephens2008}, have been achieved in other fields, suggesting that similar approaches might be possible in language as well.
Accordingly, our results should be interpreted as clarifying what internal representations look like geometrically, and what geometry implies for probing and steering, rather than as a complete account of belief formation in general settings.
\clearpage

\section*{Acknowledgments}
%
R.S. is very grateful to Chris Earls, Toni Liu, Haley Moller and Can Rager for insightful conversations and feedback on the manuscript.


\bibliography{colm2026_conference}
\bibliographystyle{colm2026_conference}

\appendix
\section*{Appendix}
\section{Convergence to the input distribution}
\label{app:convergence}
A prompt consisting of $n$ numbers contains $2n+1$ tokens: 
an initial \verb!<|begin_of_text|>! token, $n$~\textit{com2num} tokens, and $n$~\textit{num2com} tokens.
We denote by $t \geq 0$ the index of each input number, i.e. each number token has global index $2t+1$ (0-indexed).
The corresponding \textit{com2num} tokens predicting the following numbers have index $2t+2$.

\begin{figure}[h]
  \includegraphics[width=\columnwidth]{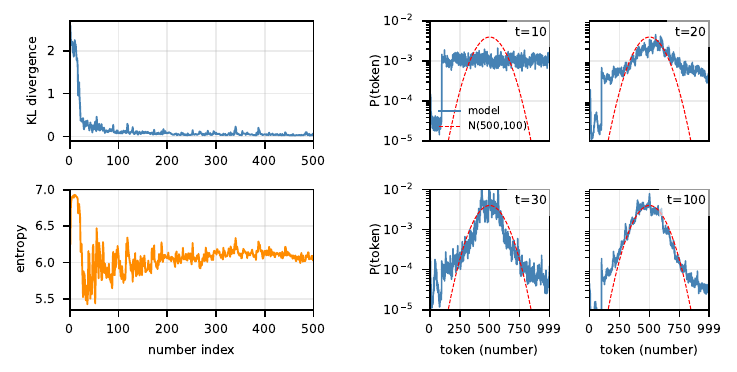}
  \caption{ 
  \textbf{Convergence to the normal distribution.} 
  (A) KL divergence between the softmax distribution and the distribution underlying the input numbers, $\mathrm{KL}(\vec{p}_t \,\|\, \mathcal{N}_{500,100})$.
  The convergence appears to be reached after about 100 numbers.
  (B) Entropy of the softmax distribution. 
  (C) Output distributions at different times, showing convergence from a uniform distribution on integers to the true distribution (red).
  }
  \label{fig:convergence}
\end{figure}

\begin{figure}[h]
  \includegraphics[width=0.5\columnwidth]{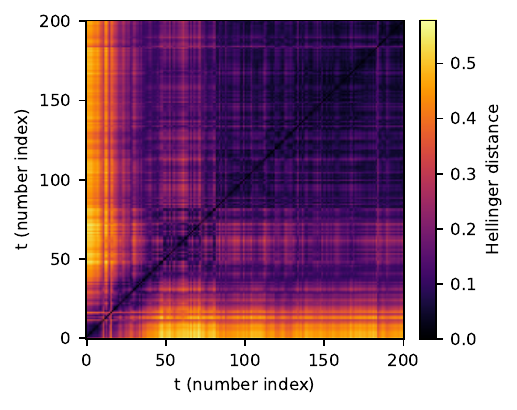}
  \caption{ 
  \textbf{Pairwise distances between output distributions.} 
  The Hellinger distance is a (symmetric) distance between two probability vectors:
  $D_\mathrm{Hel}(p_t,p_{t'}) = 2^{-1/2} \lVert \sqrt{\vec{p}_t} - \sqrt{\vec{p}_{t'}} \rVert_2$.
  Early outputs are far from those at $t \geq 100$, which encode the stationary distribution.
  }
  \label{fig:hellinger}
\end{figure}

Figure~\ref{fig:convergence} shows the convergence of the \texttt{com2num} outputs $\vec{p}_t = \vec{p}(2t+2)$ towards the true distribution underlying the data.
After about 100 numbers have been seen, the model adequately reproduces the underlying distribution in its logits.
Figure~\ref{fig:hellinger} further shows that outputs are locally close to each other, suggesting continuity of the output softmax along~$t$.

\citet{liu2025density} and \citet{bao2025texttrainedllmszeroshotextrapolate} have already studied the mechanisms of convergence in various numerical settings, starting with syntactic matching and a high-entropy phase corresponding to the uniform distribution.
We further note here that even 100 input numbers is insufficient to sketch an underlying distribution with $\sigma=100$, even though the model's output is faithful to it. 
In other words, the model \emph{believes} the input data is Gaussian, even though it doesn't yet look like it.
This suggests the model's \emph{prior} is biased towards normality.

\section{Belief manifolds for various models}
\label{app:models}
\begin{figure}[h]
  \includegraphics[width=0.92\linewidth]{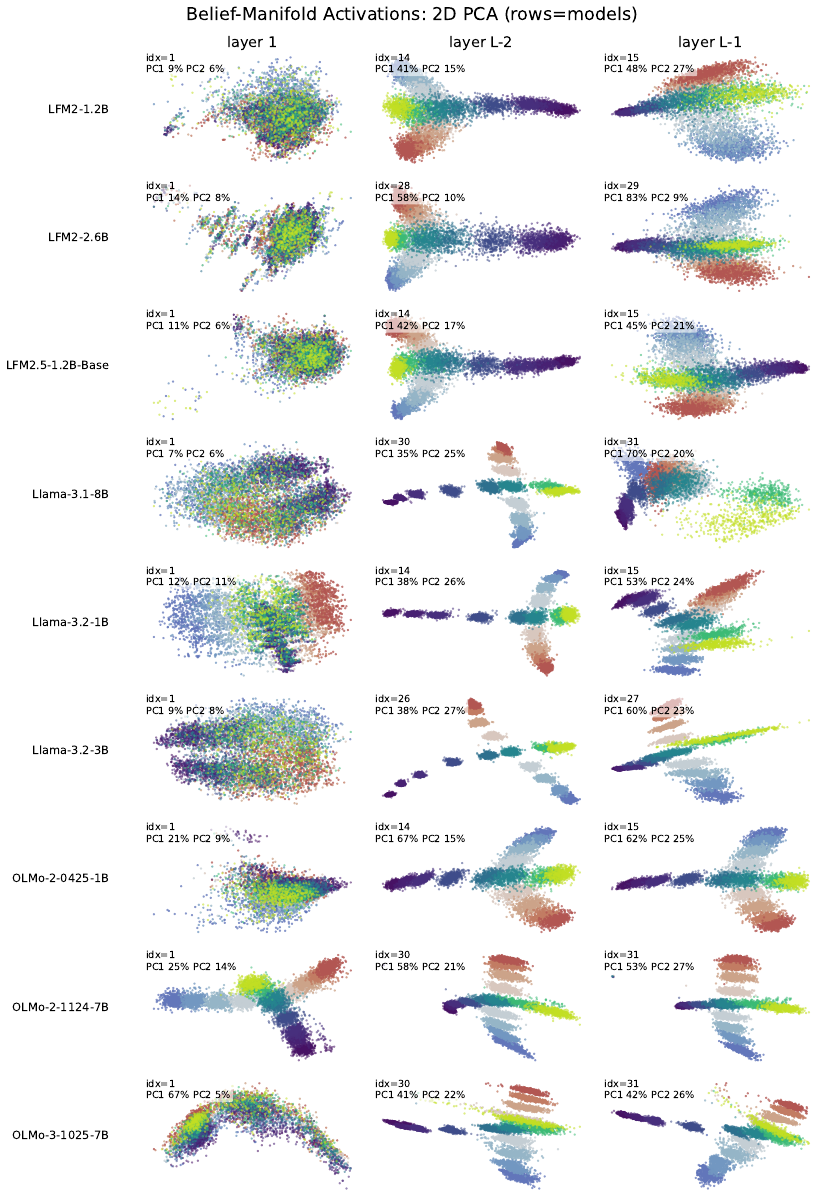}
  \caption{ 
  \textbf{Structured manifolds across models.} 
  2D PCA plots of activations from various models. $L$ is the number of layers in the model. Plotted here are residual stream activations after the second layer ($L=1$), the penultimate ($L-2$), and the last layer ($L-1$).
  }
  \label{fig:models}
\end{figure}

We apply the same methodology (input time series, PCA of representations) to the residual stream activations of various models of the \texttt{Llama}, \texttt{LFM}, and \texttt{OLMo} models. 
We focus on these models as they are the main foundational base models with unique tokens for each integer between 0 and 999.
In Fig.~\ref{fig:models}, we observe qualitatively the same structured manifolds across models, with increasing resolution as we go towards the deeper layers.
An interactive explorer of low-dimensional visualization of manifolds across models and layers will be made available.

\section{Ideal observer}
\label{sec:ideal_observer}
In a previous study, \citet{liu2025density} observed that LLMs operate an implicit, context-dependent Kernel Density Estimation procedure to compute logit distributions from a numerical input.
Here, we consider the perspective of a Bayesian observer that incorporates the sequentiality of the input data to infer its own posterior.

Considering the switching dynamics of Section~\ref{sec:belief_dynamics}, it is apparent that the language model does \emph{not} conform to an online ideal Bayesian observer under standard assumptions.
For one thing, intuitively, an ideal observer would compute a running average of the input data and hence the output distribution mean would not converge fast enough to 700 as in Fig.~\ref{fig:belief_dynamics}.

\begin{figure}[h]
  \includegraphics[width=0.5\columnwidth]{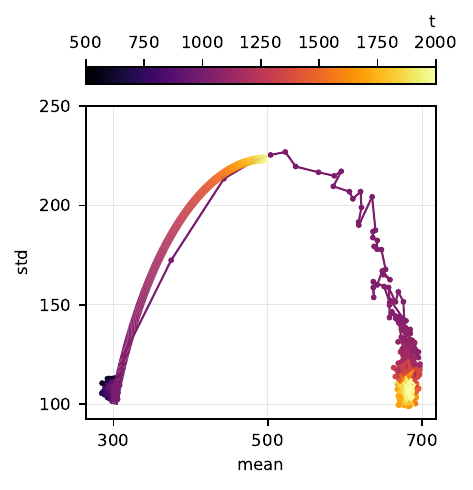}
  \caption{ 
  \textbf{LLM vs Ideal Observer.} 
    Ensemble-averaged (10 samples) trajectory of \texttt{Llama-3.2}'s softmax mean and standard deviation (line and dots), compared to the expectation from an ideal observer model, as a function of $t$ (color).
  }
  \label{fig:ideal_observer}
\end{figure}

More precisely, assuming the model's generative process is stationary \& i.i.d. Gaussian, with unknown mean and unknown variance, the output distribution should follow:
\begin{equation*}
    \sigma^2 \sim \mathrm{InvGamma}(\alpha_0,\beta_0), 
    \, \mu \mid \sigma^2 \sim \mathcal{N}\left( \mu_0, \frac{\sigma^2}{\kappa_0}\right).
\end{equation*}
The resulting posterior predictive $p(x_{t+1} | x_{1:t})$ is Student-t distributed.
The expected predictive mean reads:
\begin{equation*}
    \mathbb{E}\left[ \mathrm{avg}(x_{t+1} \mid x_{1:t}) \right] \simeq \frac{300n_1(t) + 700n_2(t)}{t}, 
\end{equation*}
where $n_1 = \min(t, 1000)$, $n_2(t) = \max(t-1000,0)$, and under the assumption of weak prior ($\kappa_0$ small), and the expected predictive standard deviation:
\begin{equation*}
    \mathbb{E}\left[ \mathrm{std}(x_{t+1} \mid x_{1:t}) \right] \simeq \sqrt{100^2 + 400^2 \frac{n_1(t) n_2(t)}{t^2}}.
\end{equation*}
In Fig.~\ref{fig:ideal_observer}, we show the trajectory of the ideal observer in the $(\mu,\sigma)$ plane on top of the ensemble-averaged trajectory of \texttt{Llama-3.2}'s softmax on the switching time series. 
The LLM converges much faster than the ideal observer.

\section{Meta-in-context learning}
\label{app:metalearning}
\begin{figure*}[t]
\centering
\begin{subfigure}{\linewidth}
\centering
\includegraphics[width=\linewidth]{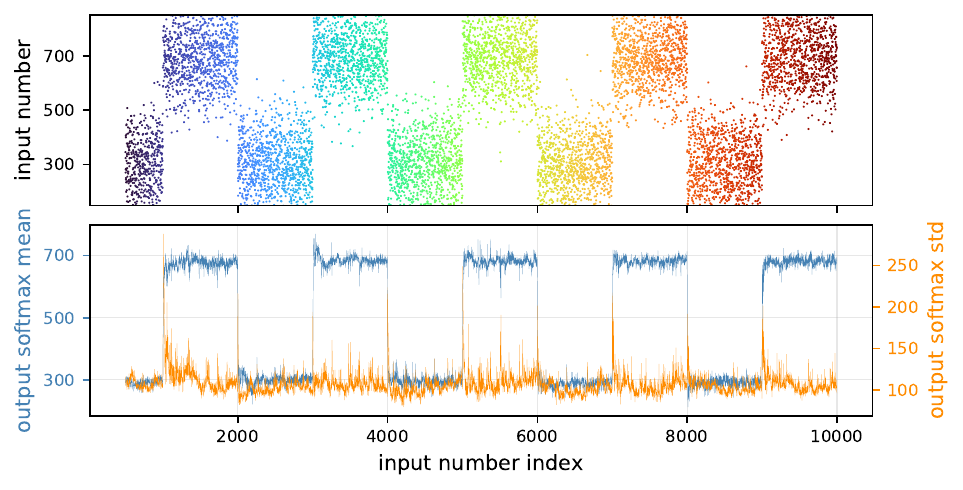}
\end{subfigure}
\begin{subfigure}{\linewidth}
\centering
\begin{overpic}[width=\linewidth]{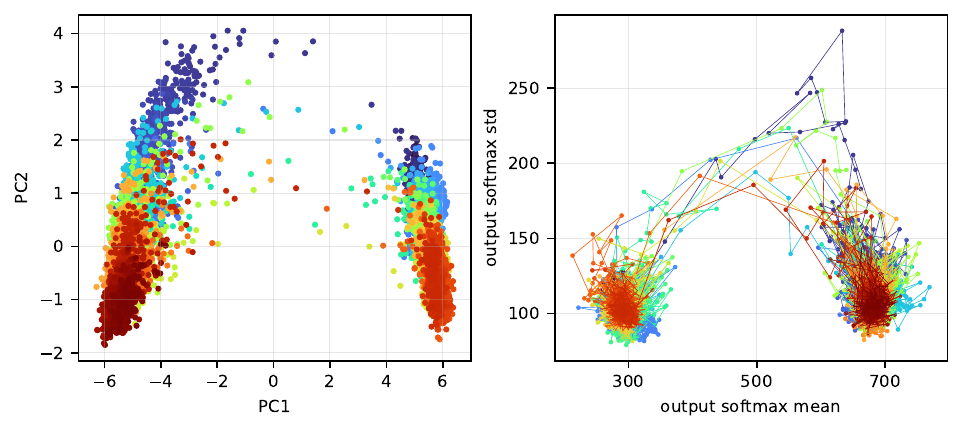}
\put(10,92){\makebox(0,0)[l]{A}}
\put(10,70){\makebox(0,0)[l]{B}}
\put(10,40){\makebox(0,0)[l]{C}}
\put(60,40){\makebox(0,0)[l]{D}}
\end{overpic}
\end{subfigure}
\caption{
  \textbf{Meta-in-context learning.}
  (A) The input time-series consists of 10 sequences of 1000~numbers each, alternating between $\mathcal{N}(300,100)$ and $\mathcal{N}(700,100)$.
  (B) Mean and standard deviation of the softmax output (next-number prediction).
  (C) Activations (layer 14) for the corresponding input tokens in (A).
  (D) Trajectory in the (mean,std) plane of the output distributions , colored by input index.
  }
\label{fig:metaICL}
\end{figure*}

In Fig.~\ref{fig:belief_dynamics}, we observed the response of the LLM upon a change of input distribution.
We now extend this experiments to a long sequence of several switches between $\mathcal{N}(300,100)$ and $\mathcal{N}(700,300)$.
The main question here is whether the model starts to understand the \emph{meta-distribution} of the input data.
Figure~\ref{fig:metaICL} provides early evidence that it does.
Indeed, the model's response to a change of distribution becomes faster and faster, as evidenced in the shape of the softmax standard deviation over time, and the trajectories of activations between the two attractors.
The model transitions much faster and follows a more direct path for later switches and the early ones.
This ability of \emph{meta-in-context learning} has been reported before, in different settings~\citep{codaforno2023metaincontextlearninglargelanguage}.

\section{Linear field probes}
\subsection{Data and feature geometry}
\label{app:dual}
\begin{figure}[h]
  \includegraphics[width=0.5\linewidth]{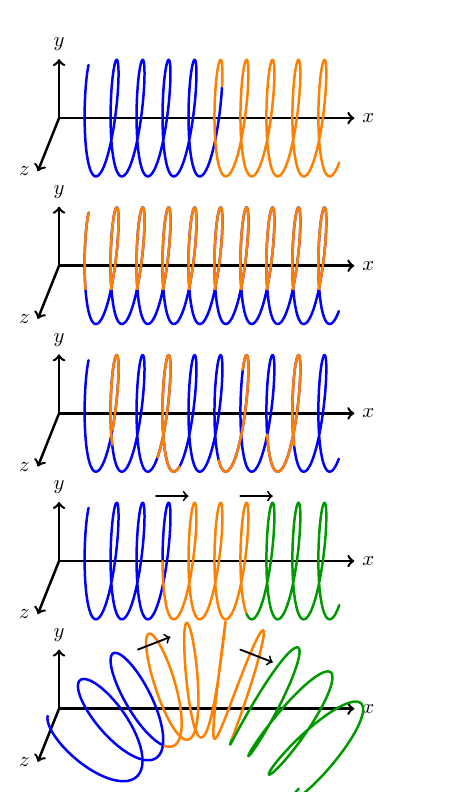}
  \caption{ 
  \textbf{Data geometry vs field geometry.} 
  (A)~The data forms a 3D helicoid, yet the two classes (blue/orange) depend only on the position along the $x$-axis: they are linearly separable (by a hyperplane perpendicular to the $x$-axis). 
  Separability does not require the data manifold to be flat.
  (B)~Another example of linearly separable classes, along the $y$-axis ($xz$-plane).
  (C)~Non-linearly separable classes: no single hyperplane separates the two colors.
  (D)~The three classes are linearly separable, with probe vectors being all collinear: the linear field probe is rank~1.
  (E)~Curved field geometry: the probe directions rotate with $\mu$, requiring at least a 2D probe subspace.
  }
  \label{fig:helicoid}
\end{figure}

In order to illustrate the dichotomy between data manifold and field geometry (primal vs dual), as framed in \citet{yocum2025neural}, we propose the visualization in Fig.~\ref{fig:helicoid}.
Suppose representations lie on a helicoid: their geometry is fully three-dimensional. 
Let's now assume that these activations describe separate classes: blue, orange, green;
class assignment varies along the manifold
There are many ways these classes can be encoded along the manifold.
In Fig~\ref{fig:helicoid}A, B, D, E these classes are \emph{linearly separable} because there exists linear probe vectors (equivalently: single hyperplanes) that separate the classes.
In Fig~\ref{fig:helicoid}C, no such linear separator exists, therefore the classes are not separable from single decision boundary directions.

Let's now consider the dimensionality of the linear field probe.
In Fig~\ref{fig:helicoid}D, the class probe vectors are collinear across $\mu$, so that $\dim \mathrm{span}\{w_\mu\} = 1$ (a rank-1 field).
However, in Fig~\ref{fig:helicoid}E, two \emph{non-parallel} hyperplanes are required, one to separate blue from orange, and another for orange to green: the span of the linear field probe is 2. 
This intrinsic field dimensionality is reflected in the rank of the LFP Gram matrix.

More intuitively, one can consider this intuitive example of the planar map of Earth.
The landmasses represent where data is distributed in space: this is the primal geometry, or data manifold.
Each point on the land is also associated with different features, for example: the country it belongs to, the local climate, or the elevation.
These features define a dual geometry: they are fields supported on the data manifold, but taking values in a different space (e.g., a finite set of countries, or $\mathbb{R}^+$ for elevation).
Some categorical features are linearly separable: Australia and New Zealand can be separated by a single line (more generally, a hyperplane).
Others are not: Belgium and the Netherlands are not cleanly separable due to intricacies and enclaves at their border~\citep{WelchLabsAI}.
Continuous feature fields, such as atmospheric pressure, have a geometry of their own, visualized by isobars.
A linear field probe corresponds to a family of linear readouts that locally approximate such feature fields, revealing their geometry through how probe directions vary across the domain.
To make this notion precise, we need to characterize how these probe vectors relate to one another globally.
Further elaboration is provided in App.~\ref{app:dual}.

\subsection{Set of linear probes}
\label{app:linearprobes}
Practically, to calculate a set of linear probes on an ensemble of activations representing a set of $C$~classes, and test the hypothesis that they constitute a LFP, we have two main options:
\paragraph{A multiclass probe,} which computes the logits:
\begin{equation*}
    \vec{z}(\vec{x}) = \vec{W} \vec{x} + \vec{b},
\end{equation*}
with $\vec{z} \in \mathbb{R}^C, \vec{W} \in \mathbb{R}^{C \times d}, \vec{b} \in \mathbb{R}^C$.
It is trained using a softmax + cross-entropy loss.
The point probes correspond to the domain embedding vectors~$\Psi(\mu_i)$ at the discretized points~$\mu_i$, i.e.~$\Psi(\mu_i)$ are the rows $\vec{w}_i$ of $\vec{W}$. 

\paragraph{A one-vs-rest probe set,} i.e. a set of $C$~binary classifiers with scores for each class~$i$:
\begin{equation*}
    s_i(\vec{x}) = \vec{w}_i^\top \vec{x} + b_i,
\end{equation*}
and corresponding probabilities $p_i(\vec{x}) = \sigma\left( s_i(\vec{x})\right)$ and a sigmoid + binary cross-entropy loss.
The point probes are the $C$ learned $\vec{w}_i \in \mathbb{R}^d$.

\paragraph{}
In either case, the bias term $b$ is optional and a weight-decay is applied for training.
We set $\vec{b}=0$ for our probes for simplicity -- it doesn't seem to affect performance.
In practice, we find one-vs-rest probes to be less efficient to separate a specific class from all others.
Hence we rely on the multiclass probe in what follows.

The steering directions are calculated from differences of point-probes.
(For multiclass probes using a softmax model, the row vectors are defined up to a shared shift, so we optionally center them across classes without affecting steering differences.)

\subsection{Manifold interpolation}
\label{sec:probe_interpolation}
We consider the unit-norm probe vectors $\vec{w}_{\mu_i}$; for simplicity let's take $\vec{w}_{300}$ and $\vec{w}_{400}$. We aim to find an interpolation for the probe vector $\vec{\tilde{w}}_{350}$ at $\mu=350$.

\paragraph{Linear interpolation.} This is simply the arithmetic weighted mean: \begin{equation*}
    \vec{\tilde{w}}_{350} = \alpha\vec{w}_{300} + (1-\alpha)\vec{w}_{400},
\end{equation*}
with $\alpha=0.5$ for the midpoint.

\paragraph{Geodesic interpolation.} We use the standard spherical interpolation formula:
\begin{equation*}
    \vec{\tilde{w}}_{350} = \frac{\sin{(1-\alpha)\theta}}{\sin{\theta}} \vec{w}_{300} + \frac{\sin{\alpha\theta}}{\sin{\theta}} \vec{w}_{400},
\end{equation*}
where $\theta = \arccos{\vec{w}^\top_{300}\vec{w}_{400}}$ and $\alpha=0.5$.

\paragraph{Kernel interpolation.} In this case, $\vec{\tilde{w}}_{350} = \sum_i a_i \vec{w}_i$, with $\vec{a} = \vec{G}^{-1} \vec{k}_{350}$ and $\vec{k}_{350} = \alpha\vec{G}_{300,j} + (1-\alpha)\vec{G}_{400,j}$, where $\vec{G}$ is the Gram matrix (the notation $\vec{G}_{300,j}$ is shorthand for $\vec{G}_{(i,j) | \mu_i = 300}$, i.e. the row of $\vec{G}$ corresponding to $\mu$=300).

\paragraph{Eigenbasis interpolation.} \citet{yocum2025neural} propose a more rigorous approach from their Field Geometry Equivalence Theorem based on spectral decomposition in feature space.

\subsection{Mixture of manifolds}
\label{app:mixture}
\begin{figure}[h]
  \includegraphics[scale=1]{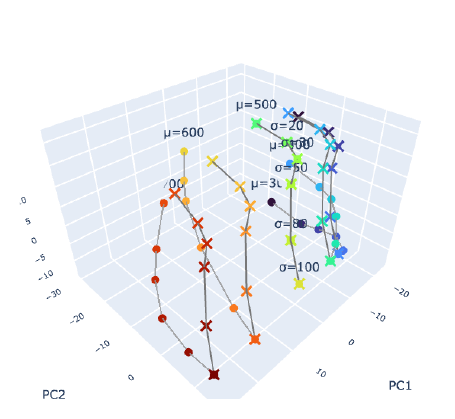}
  \caption{ 
  \textbf{Mixture of manifolds.}
  Prototypes interpolated from the anchor manifolds $\mathcal{M}_{\mu,\sigma_0}$ and $\mathcal{M}_{\mu_0,\sigma}$ with $(\mu_0=500,\sigma_0=100)$, marked with a cross, do not match the geometry of the ground-truth centroids (circles).
  }
  \label{fig:mixture_manifolds}
\end{figure}

The Linear Representation Hypothesis is a convenient framework to think of latent representations as a linear sum of features encoded as directions~\citep{elhage2022superposition, 10.5555/3692070.3693675}.
However, ample evidence has found that features are often encoded along curved manifolds, such as circular supports~\citep{gurnee2025when}.
Our previous observations herein concur.
This yields the question: given manifold parametrizations, can we instead interpolate any combination of parameters as a \textit{mixture of manifolds}?
More precisely, around an anchor $(\mu_0,\sigma_0)$, we hypothesize that the effect of $\mu$ and $\sigma$ on the representations decomposes into independent subspaces $U$ and $V$ such that
\begin{equation*}
    c(\mu,\sigma) \simeq c(\mu_0,\sigma_0) + u(\mu) + v(\sigma),
\end{equation*}
with $u(\mu) \in U, v(\sigma) \in V$, and the gauge fixed at $u(\mu_0)=v(\sigma_0)=0$.
Under this model, the surface of prototypes is approximately a product manifold embedded as an additive superposition.
We investigate experimentally by fitting a spline $c_\mu(\mu)$ through the centroids of the $\mathcal{M}_{\mu,100}$ tiles and another spline $c_\sigma(\sigma)$ across $\mathcal{M}_{500,\sigma}$.
Then, we interpolate:
\begin{align}
    \tilde{c}(\mu^\ast,\sigma^\ast) = c_0 &+ \left( c_\mu(\mu^\ast)-c_\mu(\mu_0) \right) \notag \\
    &+ \left( c_\sigma(\sigma^\ast)-c_\sigma(\sigma_0) \right), \notag
\end{align}
with $c_0 = c(\mu_0,\sigma_0)$ denoting an anchor point, for example $(\mu=500,\sigma=100)$.
Fig.~\ref{fig:mixture_manifolds} shows the interpolated centroids compared to the ground-truth centroids computed from the activations of the corresponding time series.
Evidently, the simple, spline-based interpolation fails to capture the true geometry of the $(\mu,\sigma)$ sheet.
This is possibly due to non-linear interactions between parameters, here $\mu$ and $\sigma$.
Other decompositions might be possible.
For example, \citet{fel2025rabbithulltaskrelevantconcepts} recently proposed the Minkowski Representation Hypothesis, according to which concepts are represented as convex hulls, and representations can de decomposed as (non-unique) sums of polytopes.

\end{document}